\documentclass[letterpaper, 10 pt, conference]{ieeeconf}

\IEEEoverridecommandlockouts 
\usepackage{tabularx, graphicx,
  amsmath,amsfonts,siunitx,pifont,amssymb,subcaption}
\usepackage{xcolor, bm, float, multirow, multicol, courier, mathtools, url}
\usepackage{footmisc}
\usepackage[linesnumbered,ruled,vlined]{algorithm2e} \pdfminorversion=4
\let\labelindent\relax \usepackage{enumitem} \usepackage{graphicx}

\usepackage{siunitx}

\usepackage{pdfpages}
\usepackage{cite}
\usepackage{algpseudocode}
\usepackage{float}
\usepackage{colortbl}
\usepackage{dblfloatfix}    
\usepackage{balance}

\def\secref#1{Sec.~\ref{#1}}
\def\figref#1{Fig.~\ref{#1}}
\def\tabref#1{Tab.~\ref{#1}}
\def\eqref#1{Eq.~(\ref{#1})}

\title{\LARGE \bf The Newer College Dataset:\\Handheld LiDAR, Inertial and Vision with Ground Truth}

\author{Milad Ramezani, Yiduo Wang, Marco Camurri, David Wisth, Matias Mattamala and Maurice
Fallon
	\thanks{The authors are with the Oxford Robotics Institute, University of Oxford, UK.
		{\tt\small \{milad, ywang, mcamurri, davidw, matias, mfallon\}@robots.ox.ac.uk}}%
}

\begin{document}

\setlength{\abovedisplayskip}{4pt}
\setlength{\belowdisplayskip}{4pt}

\maketitle \thispagestyle{empty} \pagestyle{empty}

\begin{abstract}
In this paper, we present a large dataset with a variety of mobile mapping
sensors collected using a
handheld device carried at typical walking speeds for nearly 2.2 km
around New College, Oxford as well as a series of supplementary datasets with much more aggressive
motion and lighting contrast.
The datasets include data from two commercially available devices - a stereoscopic-inertial camera
and a multi-beam 3D LiDAR, which also provides inertial measurements. Additionally, we used a
tripod-mounted survey
grade LiDAR scanner to capture a detailed millimeter-accurate 3D map of the test location
(containing $\sim$290 million points). Using the map, we generated a 6
Degrees of Freedom (DoF) ground truth pose for each
LiDAR scan (with approximately 3~cm accuracy) to enable
better benchmarking of
LiDAR and vision localisation, mapping and reconstruction systems. This ground truth is the
particular novel contribution of this dataset and we believe that it will enable systematic
evaluation which many similar datasets have lacked. The large dataset combines both built
environments,
open spaces and vegetated areas so as to test localisation and mapping systems such as vision-based
navigation, visual and LiDAR SLAM, 3D LiDAR reconstruction and appearance-based place recognition,
while the supplementary datasets contain very dynamic motions
to introduce more challenges for visual-inertial odometry systems. The
datasets are available at:\\ \url{ori.ox.ac.uk/datasets/newer-college-dataset}
\end{abstract}


\begin{figure}
\includegraphics[width=\linewidth]{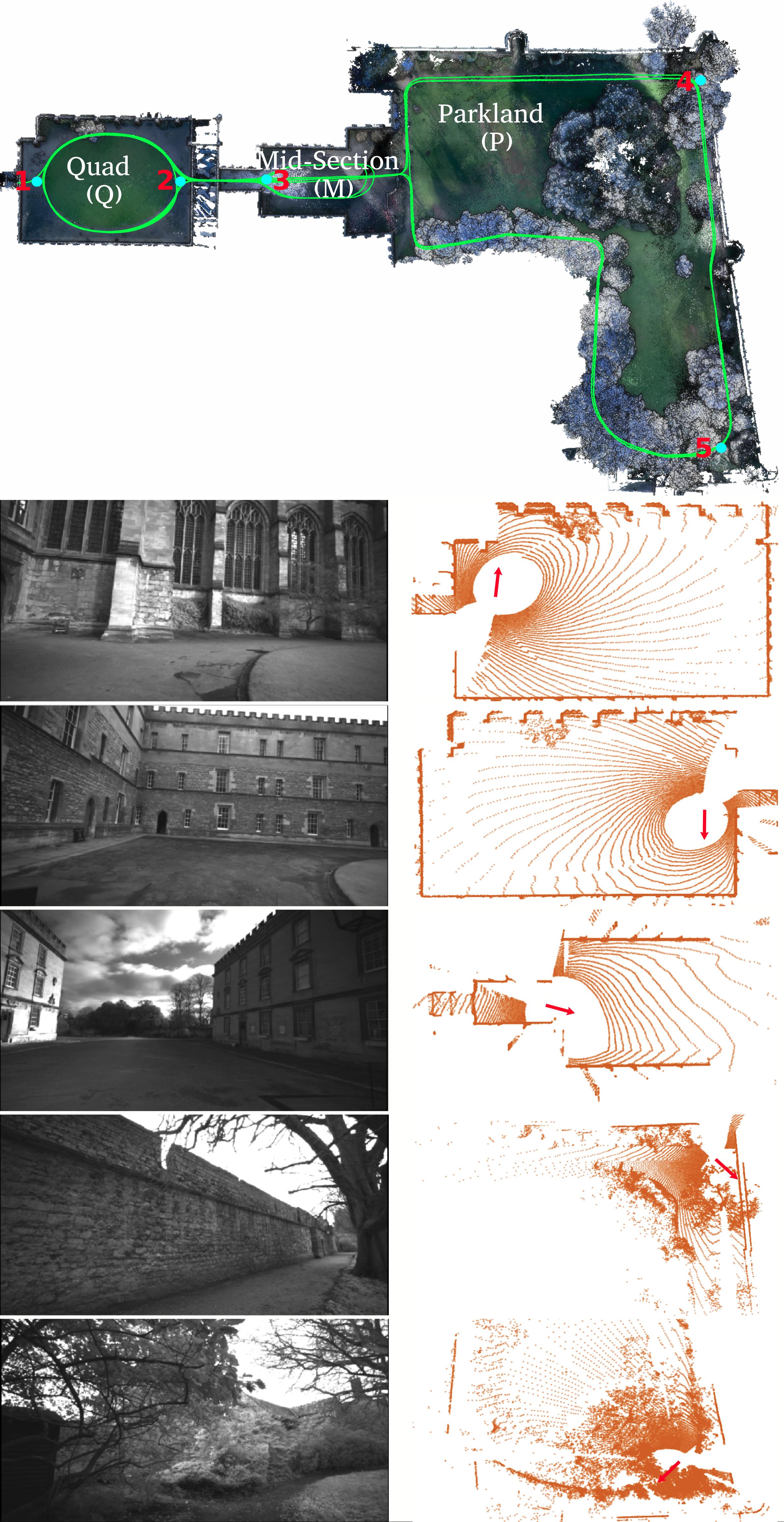}
\caption{\small{\textbf{Top}: A bird's eye view of the 3D model of New College generated with
a Leica BLK360 scanner and the ground truth trajectory obtained by the proposed approach.
\textbf{Bottom}: Image samples and corresponding 3D LiDAR scans from
the Quad (locations 1 and 2), the Mid-Section (location 3) and the Parkland
(location 4 and 5) show varied environments suited for assessment of localisation
and mapping algorithms in autonomous robot navigation.}}
\label{fig:teaser}
\end{figure}


\section{Introduction}
Research in robotics and autonomous navigation has benefited significantly from the public
availability of standard datasets which enable systematic testing and validation of algorithms.
Over the past 10 years, datasets such as KITTI~\cite{Geiger2013IJRR}, the New College
\cite{smith2009new} and EuRoC MAV \cite{Burri25012016} have been released and provided a transparent
benchmark of performance. These datasets were collected on a variety of platforms (UGVs, UAVs and
Autonomous Vehicles) with reference poses based upon GPS/INS
\cite{Geiger2013IJRR}, dead reckoning \cite{smith2009new}, and laser
trackers/motion capture systems~\cite{Burri25012016}.

Many vehicular benchmarks use a form of GPS/INS fusion for ground truth but do not provide
precise local accuracy. This is specially noted as a shortcoming of the KITTI odometry benchmark which does
not use length scales of less than
100~m for this reason\footnote{http://www.cvlibs.net/datasets/kitti/eval\_odometry.php}. Tripod-mounted laser trackers
follow a prism placed on the robot to achieve precise localisation. However,
they cannot maintain
line of sight to the robot for large
experiments. Motion capture systems provide accurate 6 DoF ground truth but are limited to small
indoor facilities.

We aim to provide a large-scale dataset which provides centimeter accuracy so as to enable
evaluation of short length-scale odometry, as well as large scale drift. The raw data files are
accompanied by a precise 3D model constructed using a survey-grade LiDAR scanner. Utilizing the 3D
model,
we inferred the location of the device using LiDAR ICP at 10 Hz across the entire
location.~\figref{fig:teaser} shows a plan view of the 3D model with vision and corresponding LiDAR
samples of the data from different locations.

The data was collected by a handheld device carried by a person at walking speed, unlike the
majority of the released datasets acquired from a robotics platforms. The handheld device comprises
a 3D LiDAR and a stereo camera each with a self-contained IMU. In particular, we
use commercially
available low-cost sensors --- the widely used Intel Realsense D435i and a
64-beam Ouster OS-1 LiDAR
scanner. The walking motion can, to a degree, replicate the jerky motion of a flying drone or a quadruped.

This dataset revisits New College, Oxford --- the location of the original Smith \textit{et al.}~\cite{smith2009new} 
dataset. This dataset is particularly useful due to its variety of
scenarios, including structured buildings, vegetation, and open space
areas with lack of texture. We replicate the sequences of the original dataset
and then go on to add more aggressive, faster sequences designed to 
test algorithms for visual navigation, LiDAR SLAM, reconstruction and place recognition.

The remainder of the paper is structured as follows: \secref{sec:relatedWorks} reviews related work
followed by a description of the device in \secref{sec:platform}.~\secref{sec:dataCollection}
details our dataset.~\secref{sec:groundTruth} explains how we determined the ground
truth.~\secref{sec:dataset-usage} demonstrates example usage of our dataset for a set of navigation
research topics in mobile robotics before a summary is presented
in~\secref{sec:conclusion}.


\section{Related Work}
\label{sec:relatedWorks}
Dataset papers can be divided into two subsections based on the platform carrying the sensor; a
(self-driving) car is often used for outdoor, large-scale datasets while 
robots or humans-carried datasets are typically much smaller and often indoors.
Focusing on vision, IMU and LiDAR modules,~\tabref{table:AteRunTimeStats}, provides the basic
details of the datasets discussed in the following.

\subsection{Vehicle-based Datasets}

There is a large body of localisation and mapping data targeting autonomous navigation for ground
vehicles.

The MIT DARPA Urban Challenge dataset~\cite{huang2010high} is one of the first
major UGV datasets. It
was collected
using MIT's Talos vehicle over the course of a 90 km traverse spanning 7 hours of
self-driving. The ground truth was provided by the integration of a high-precision GPS and an INS.

The Marulan multi-modal
datasets~\cite{peynot2010marulan} were gathered by an unmanned ground vehicle, Argo, in which
artificial dust, smoke and rain added extra challenges for on-board perception sensors.

Similarly to the MIT dataset, the Ford Campus dataset~\cite{pandey2011ford} was obtained
along an almost 6 km traverse of a mock-up urban environment. The Malaga urban
dataset~\cite{blanco2014malaga} has the distinctive feature of high-resolution pair images captured for
over an almost 37 km trajectory in urban scenarios.

The KITTI dataset~\cite{Geiger2013IJRR} was collected on a car driving around the streets of
Karlsruhe and it has significantly catalyzed autonomous vehicle navigation research. The dataset consists of
stereoscopic image pairs for the sequences of 39.2 km length and has been used for a variety of
SLAM/odometry and
object detection tasks. KITTI provides 6 DoF ground truth trajectory for all the traversals using
RTK-GPS/INS with accuracy below 10 cm. However, this accuracy is not guaranteed in GPS-deprived
areas such as urban canyons. Furthermore, the IMU readings and images are not synchronized which
effects the performance of many visual-inertial odometry algorithms.

The longest autonomous driving dataset we are familiar with is the Oxford RobotCar dataset~\cite{maddern20171} collected in all
natural weather conditions over the course of 1000 km driven through central Oxford. Recently,
the Complex Urban LiDAR dataset~\cite{jeong2019complex} was gathered and is targeted at multiple challenges in
complex urban areas including GPS loss, multi-lane highways and dynamic entities such as
pedestrians, bikes and cars. This multi-faceted dataset was acquired over the course of
approximately
180~km of travel.

Nonetheless, the ground truth of the aforementioned driverless-car datasets is highly dependent
on GPS observations and therefore, as noted in~\cite{maddern20171} and~\cite{jeong2019complex},
the usage of ground truth is not recommended in GPS-deprived areas for the evaluation of
localisation and mapping algorithms.

\subsection{Mobile Robot or Human-carried Datasets}

The New College Vision and LiDAR dataset~\cite{smith2009new}, which is a motivation for this
dataset, provides carefully timestamped laser range data, stereo and omnidirectional imagery
along with 5 DoF odometry (2D position and roll, pitch, heading).
The data was collected using a wheeled robot, a Segway, over a 2.2 km traverse of the college's
grounds and the adjoining garden area. No ground truth is available for this dataset.


\begin{table*}[t]
	\centering
	\resizebox{\linewidth}{!}{%
	\begin{tabular}{llllllll}
		\hline
		 \multirow{2}{*}{\textbf{Dataset}}  &
         \multirow{2}{*}{\textbf{Year}} &
         \multirow{2}{*}{\textbf{Environment}} &
         \multirow{2}{*}{\textbf{Ground Truth}}&
         \multicolumn{3}{c}{\textbf{Sensors}}&
         \multirow{2}{*}{\textbf{Platform}}\\
		 \cline{5-7}
		\multirow{-2}{*}{} & & &
&\textbf{IMUs}&\textbf{LiDAR}
&\textbf{Cameras}&\\

        \hline \hline
		\textbf{Rawseeds~\cite{ceriani2009rawseeds}} &2009 & Structured & 2D+Yaw&accel/gyro @128Hz &
2
         2D-Hokuyo @10Hz &Trinocular Vision: 3$\times$640$\times$480 @30Hz&Wheeled  \\
		 & &  & Visual Markers/Laser& &2 2D-SICK @75Hz & RGB: 640$\times$480 @30Hz&Robot\\
		 & &  & & &  & Fisheye RGB: 640$\times$640 @15Hz\\
		\hline
		\textbf{New College~\cite{smith2009new}} & 2009 & Structured
		&N/A &gyro @28Hz  &2 2D-SICK @75Hz &BumbleBee: 2$\times$512$\times$384 @20Hz&Wheeled \\
		& & Vegetated
		& & & &LadyBug 2: 5$\times$384$\times$512 @3Hz&Robot\\
		\hline
		\textbf{DARPA~\cite{huang2010high}} & 2010 &Structured &GPS/INS &N/A &12 2D-SICK @
        75Hz& 4 Point Grey: 4$\times$376$\times$240 @10Hz &Car \\
		& & Urban& & &3D-Velodyne HDL-64E @15Hz &Point Grey: 752$\times480$ @22.8Hz, Wider FOV  \\
		\hline
        \textbf{Marulan~\cite{peynot2010marulan}} & 2010 &Open Area & DGPS/INS &accel/gyro @50Hz & 4
        2D-SICK @18Hz &Mono Prosilica: 1360$\times$1024 @10Hz &Wheeled \\
          &  & & & & &Infrared Raytheon: 640$\times$480 @12.5Hz &Robot\\
		\hline
		\textbf{Ford Campus~\cite{pandey2011ford}} & 2011 &Urban &GPS/INS & accel/gyro @100Hz&
3D-Velodyne
        HDL-64E @10Hz&LadyBug 3: 6$\times$1600$\times$600 @8Hz&Car \\
		 & & & & &2 2D-Riegl LMS @& \\
		\hline
		\textbf{KITTI~\cite{Geiger2013IJRR}} & 2013 &Structured &RTK GPS/INS & accel/gyro @10Hz&
        3D-Velodyne HDL-64E @10Hz & 2 Point Grey(gray): 2$\times$1392$\times$512 @10Hz& Car\\
		&  &Urban & & & &2 Point Grey(color): 2$\times$1392$\times$512 @10Hz \\
		\hline
		\textbf{Malaga~\cite{blanco2014malaga}} & 2014 &Structured &N/A &accel/gyro @100Hz &3
        2D-Hokuyo @40Hz &BumbleBee: 2$\times$1024$\times$768 @20Hz&Car \\
		&  & Urban& & &2 2D-SICK @75Hz & \\
		\hline
		\textbf{NCLT~\cite{carlevaris2016university}} & 2015 &Structured &RTK-GPS &accel/gyro @100Hz
         & 3D-Velodyne HDL-32E @10Hz&LadyBug 3: 6$\times$1600$\times$1200 @5Hz &Wheeled \\
		 &  &Urban &LiDAR-SLAM & &2 2D-Hokuyo @10/40HZ & &Robot\\
		\hline
		\textbf{EuRoC MAV~\cite{burri2016euroc}} & 2016 &Structured &6DOF Vicon &accel/gyro @200Hz &
N/A&2
         MT9V034: 2$\times$752$\times$480 @20Hz& UAV \\
		 &  & & 3D Laser Tracker& & & \\
		\hline
		\textbf{PennCOSYVIO~\cite{pfrommer2017penncosyvio}} & 2017 &Structured &Visual Tags &
        ADIS accel/gyro @200Hz &N/A &3 GoPro (color): 3$\times$1920$\times$1080 @30Hz \\
		 &  & & &2 Tango accel @128Hz & &2 MT9V034 (gray): 2$\times$752$\times$480 @20Hz&Handheld \\
		 &  & & &2 Tango gyro @100Hz & & \\
		\hline
		\textbf{Zurich Urban MAV~\cite{majdik2017zurich}} & 2017 &Structured &Aerial  &
        accel/gyro @10Hz &N/A &GoPro (color): 1920$\times$1080 @30Hz&UAV \\
		 &  &Urban & Photogrammetry& & & \\
		\hline
		\textbf{Oxford RobotCar~\cite{maddern20171}} & 2017 &Structured &GPS/INS &accel/gyro @50Hz
        &2 2D-SICK @50Hz & BumbleBee: 2$\times$1280$\times$960 @16Hz&Car\\
		 &  &Urban &Not Recommended & &3D-SICK @12.5Hz &3 Grasshoper2:
3$\times$1024$\times$1024
         @11.1Hz \\
		\hline
		\textbf{TUM VI~\cite{schuberttum}} & 2018 &Structured &6DOF MoCap  &accel/gyro @200Hz &
N/A&IDS
        (gray): 2$\times$1024$\times$1024 @20Hz &Handheld \\
		 &  & & Available at Start/End& & & \\
		\hline
		\textbf{Complex Urban~\cite{jeong2019complex}} & 2019 &Structured &SLAM &accel/gyro
         @200Hz &2 3D-Velodyne-16 @10Hz &FLIR (color): 2$\times$1280$\times$560 @10Hz& Car \\
		 &  &Urban &Not Recommended &FOG @1000Hz &2 2D-SICK @100Hz & \\
		\hline
		\textbf{Our Dataset} &2020 &Structured &6DOF ICP  &accel @250Hz
&3D-Ouster-64 @10Hz &
        D435i (Infrared): 2$\times$848$\times$480 @30Hz& Handheld \\
		 &  & Vegetated & Localisation &  gyro
@400Hz & & \\
		\hline
	\end{tabular}
	}
	\caption{ \small{Comparison of related datasets used in robotics and autonomous systems
    research. }}
	\label{table:AteRunTimeStats}
	\vspace{-1em}
\end{table*}

Similar to the New College, the North Campus Long-Term (NCLT)
dataset~\cite{carlevaris2016university} was
gathered across a college campus, indoor and outdoor, over 147.4 km traverse and 15 months, again with a
Segway. The significant difference is the provision of the ground truth using LiDAR scan matching
and high-accuracy RTK-GPS. Although this approach potentially provides centimeter accuracy, it is
susceptible
to drift indoors or near buildings which cause GPS multi-path errors.

Recent datasets such as EuRoC MAV~\cite{Burri25012016}, Zurich Urban MAV~\cite{majdik2017zurich}
and PennCOSYVIO~\cite{pfrommer2017penncosyvio} specifically focused on visual-inertial
odometry and visual SLAM. The data in EuRoC and Zurich was gathered using a
micro aerial vehicle flying indoor and outdoor for 0.9 km and 2 km, respectively, while
the data in PennCOSYVIO was obtained from a handheld device, similar to our platform, along
0.6 km trajectory outdoor. 

To provide accurate ground truth at a millimeter level, EuRoC MAV employed a laser tracker
and a motion capture system. However, a laser tracker only provides measurement of position but
not orientation. Additionally, tracking is lost if the robot travels beyond the
line of sight of the tracker. Motion capture systems are limited to the experiments within small
areas and indoors. The ground truth in Zurich Urban and PennCOSYVIO was obtained using
aerial photogrammetry and close-range photogrammetry, respectively. Nonetheless, because
photogrammetric techniques rely on image observations, it is hard to achieve an
accuracy below 10 cm, as reported in~\cite{pfrommer2017penncosyvio}, if the observations are
not within a few meters from the camera.     


The Rawseeds dataset~\cite{ceriani2009rawseeds} was used to develop vision and LiDAR-based
techniques for indoor navigation. By deploying multiple pre-calibrated cameras
or laser scanners in the operating environments, an external network is formed from which
the trajectory of the robot was estimated. However, these techniques require continuous line of
sight limiting the scale of experiments.

We use a unique approach for determining ground truth that, to the best of our knowledge,
has not been used in the published datasets. Our approach is based upon the registration
of individual LiDAR scans with an accurate prior map, utilizing ICP. The method is properly
explained in~\secref{sec:groundTruth}.

\section{The Handheld Device}
\begin{figure}[t]
\centering
  \begin{minipage}{.5\textwidth}
  \centering
  \resizebox{0.9\linewidth}{!}{
   \includegraphics[width=\linewidth]{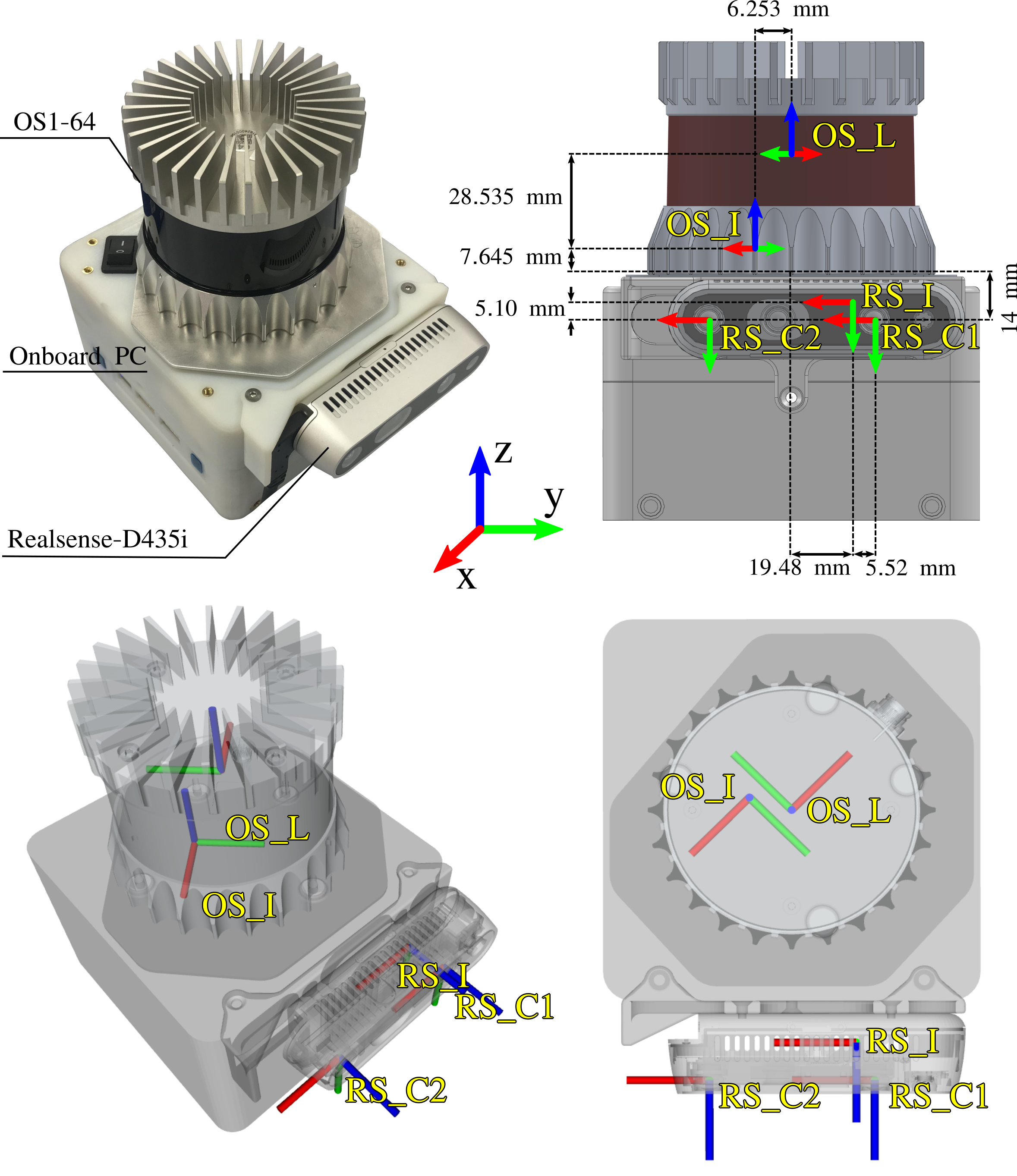}}
   \vspace{5mm}
  \end{minipage}
  \begin{minipage}{.5\textwidth}
 \centering
   \includegraphics[width=0.9\linewidth,trim={0cm 0cm 0cm
0cm},clip]{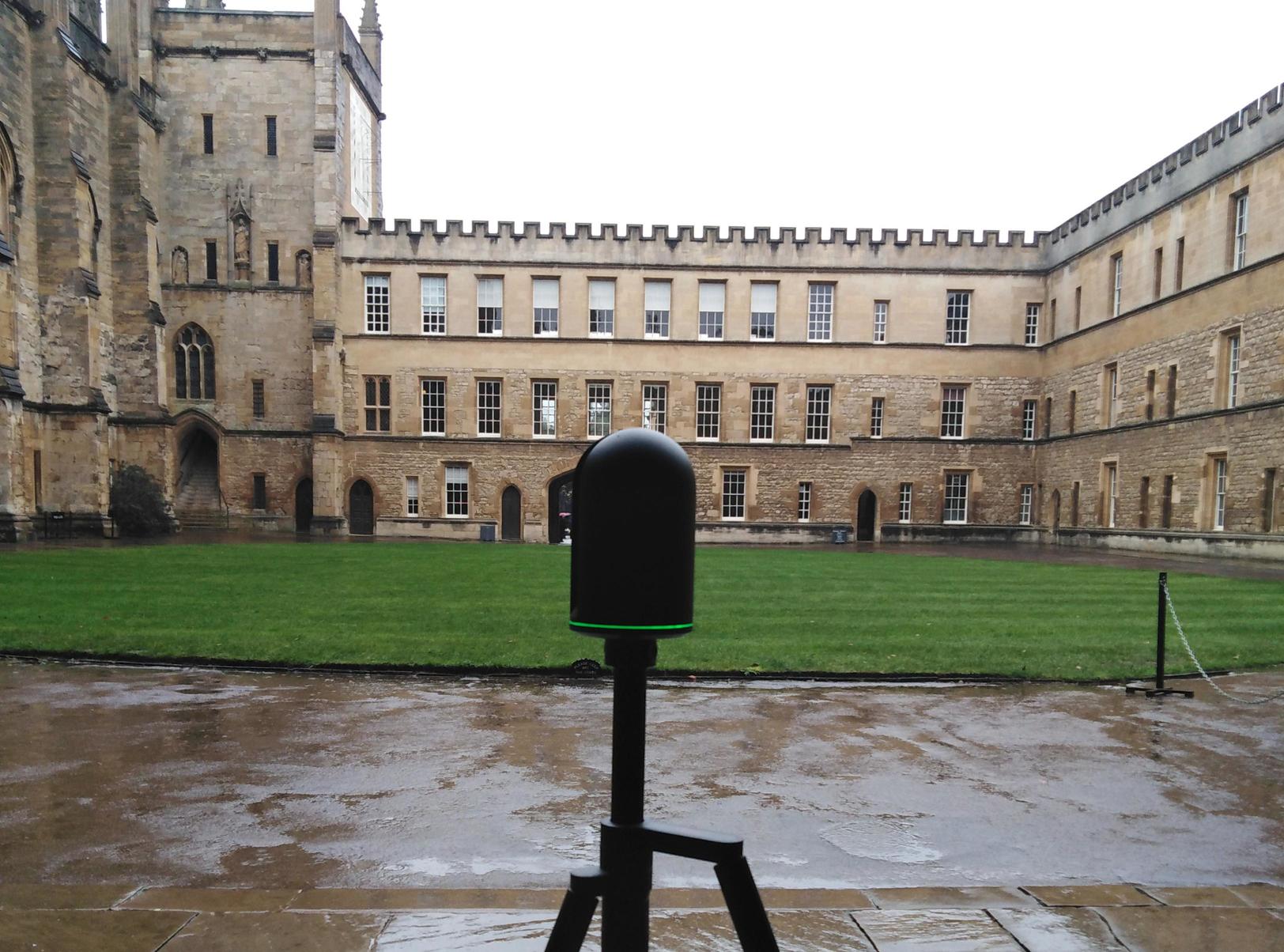}\\
   \vspace{0mm}
  \end{minipage}
     \caption{\small{Devices used in our dataset. \textbf{Top left:} Our custom
built handheld device; \textbf{Top right:} front view of the 3D CAD model with
reference frames their relative distances; \textbf{Middle (left to right):}
Isometric and top down views of the URDF model with reference frames;
\textbf{Bottom:} Leica BLK360 for creating the ground truth map.}}
\label{fig:rooster-blk}
\vspace{-5mm}
\end{figure}

Our device is shown in~\figref{fig:rooster-blk} (top-left). The sensors are rigidly attached to a
precisely 3D-printed base. The top-right figure shows a 3D model of the device from
front view. A complete URDF model of the device is available as open
source ROS
package\footnote{{https://github.com/ori-drs/rooster\_description}} and it
is shown in \figref{fig:rooster-blk} (middle). The Ouster LiDAR is mounted on
the top, with a clockwise rotation of 45 degrees for cable routing.
\tabref{table:sensorsOverview} overviews the sensors used in our handheld
device.

The Intel Realsense is a commodity-grade stereo-inertial camera while the Ouster LiDAR has 64 beams,
which provides much denser data than many other LiDAR datasets. Both sensors have become commonly
used in mobile
robotics in the last 2 years, for example the ongoing DARPA Subterranean Challenge.
\begin{table}
 \centering
 \resizebox{\linewidth}{!}{%
 \begin{tabular}{llll}
  \hline
  \textbf{Sensor}&\textbf{Type}&
  \textbf{Rate}&\textbf{Characteristics}\\
 \hline
 \hline
 LiDAR&Ouster, OS1-64 &10 Hz &64 Channels, 120 m Range\\
  & & &45$^{\circ}$ Vertical FOV\\
  & & &1024 Horizontal Resolution\\
 Cameras&Intel Realsense-D435i &30 Hz &Global shutter (Infrared)\\
  & & &848$\times$480\\
 LiDAR IMU&ICM-20948 &100 Hz &3-axis Gyroscope\\
 & & &3-axis Accelerometer\\
 Camera IMU&Bosch BMI055 &400 Hz &3-axis Gyroscope\\
 & &250 Hz &3-axis Accelerometer\\
 \hline
 \end{tabular}
 }
 \caption{\small{Overview of the sensors in our handheld device.}}
 \label{table:sensorsOverview}
 \vspace{-2mm}
\end{table}

To distinguish the sensor frames, we use the following abbreviations:
\begin{itemize}
\item OS{\_}I: The IMU coordinate system in the LiDAR.
\item OS{\_}L: The LiDAR coordinate system with respect to which the point clouds are read.
\item RS{\_}C1: The left camera coordinate system which is considered as the base frame.
\item RS{\_}I: The IMU coordinate system in the stereo setup.
\item RS{\_}C2: The right camera coordinate system.
\end{itemize}

We use the open source camera and IMU calibration toolbox
Kalibr~\cite{furgale2013unified,rehder2016extending} to compute the intrinsic calibration of the
Realsense cameras as well as their extrinsics. As our device is not hardware
synchronized, it is crucial to leverage as much software/network synchronization as possible. We
perform spatio-temporal calibration between the
cameras and the two IMUs embedded in the Realsense and the Ouster sensor. As
described in~\cite{furgale2013unified}, the temporal offsets between measurements of the Realsense
IMU and
Ouster IMU with respect to the Realsense cameras are estimated using batch, continuous-time,
maximum-likelihood estimation. In addition, by comparing the angular velocities of both IMUs, we
found out the sensors drift relative to one another at about 58 ms per hour. We studied it and
details are presented on the dataset website.
We also provide the calibration dataset along with the main dataset.
The Ouster LiDAR synchronizes with the recording computer using the Precision Time Protocol (PTP),
which achieves sub-microsecond accuracy~\cite{ptp}.

Our handheld device is equipped with an onboard Intel Core i7 NUC computer kit. To get the correct
timestamps for IMU messages of the Realsense D435i, we use Ubuntu 18.04 with the
Linux kernel of
4.15.0-74-generic and version 2.32.1.0 for Realsense
libraries\footnote{https://github.com/IntelRealSense/librealsense}. The firmware installed on the
sensor was version 0.5.10.13.00. In our experience, we found the Realsense to be sensitive to the
right combination of kernel, driver and firmwares to be used. In particular
for the IMU messages, special care needs to be taken to achieve a
reliable configuration. The device was powered by an 8000 mAh LI-PO battery.


\label{sec:platform}

\section{Data Collection}
\label{sec:dataCollection}
We have collected a variety of datasets with different speeds of walking and turning.
They are organised by the aggressiveness of the motion and described in more detail on our website. 

Since this paper is motivated by the New College dataset~\cite{smith2009new}, the longest experiment 
carefully followed the same path that the original data collection followed in 2009 in New
College, Oxford, UK. Borrowing the terminology from \cite{smith2009new}, we break the dataset into 3
main sections: Quad (\textbf{Q}), Mid-Section (\textbf{M}) and Parkland (\textbf{P}). Quad has an
oval lawn area at the center and is surrounded with medieval buildings with repeating architecture.
The Mid-Section includes a short tunnel where illumination changes quickly,
leading to an open area which is flanked by buildings on the northern and southern sides.
Parkland is a garden area connected to the Mid-Section through a wrought iron gate from west.

The data was gathered during early February 2020 from morning to noon. The handheld device was held
by a person walking at constant pace about 1 m/s. To reduce the number of blocked laser beams, the
device was held above the shoulder throughout the dataset. It is worth
mentioning that the movement was not intended to be highly dynamic. However,
natural vibration caused
by human walking and hand motion is inevitable. The motion induced by this walking gait would 
make the dataset similar to a flying UAV.

\begin{figure}[t]
\centering
\includegraphics[width=1\linewidth,trim={0cm 0cm 0cm
0cm},clip]{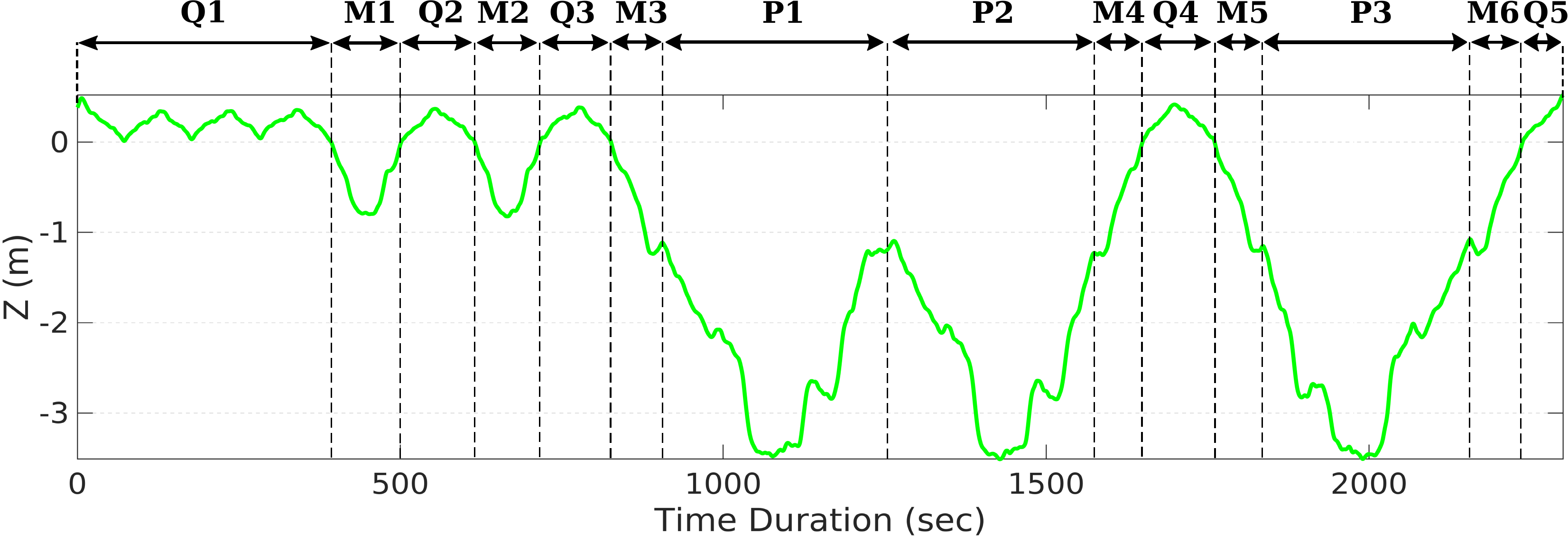}
\caption{\small{Ground truth elevation in the entire dataset.}}
\label{fig:gt-z}
\vspace{-5mm}
\end{figure}

Following the same path as the original New College dataset, the data collection began from the west
of the Quad. As illustrated in~\figref{fig:gt-z}, after three and a half loops, clockwise, around
the Quad with the duration of about 390 seconds (Q1), the Quad and the Mid-Section were traversed
back and forth twice (M1-Q2 and M2-Q3), counter-clockwise in periods M1, M2 and Q2 while clockwise
in period Q3. These sections took until second 820 of the data collection, followed by a straight
traversal in period M3 which took 60 seconds.

The data collection continued by entering the Parkland at second 1240 and it was circumnavigated
twice clockwise in 610 seconds. Unlike the paved path in the Quad and the Mid-Section, the path in
the Parkland was gravel and was muddy in parts due to the time of data collection.
Since the path is adjacent to a vegetated border and partly passes along dense foliage, it is hard
to see any building structure, posing a challenge to vision-based localisation
techniques.

The data continued to be captured by walking straight back to the Quad (M4 with the same duration
as M3) and this time the sensor was carried counter-clockwise for about 105
seconds, followed by walking
straight back to the Parkland and taking in an extra loop in this area
(but counter-clockwise). The
traversal in the P3 ended at Second 2180. Finally, the data collection ended by walking back to
starting point, i.e. passing through the Mid-Section (M6) and half circumnavigating the Quad (Q5)
counter-clockwise. Altogether, the time duration of the entire dataset is 2300 seconds.

To extend the usefulness of the dataset, we carried out further experiments with faster walking and aggressive motions of the device. These datasets are listed as below:

\begin{itemize}
 \item \textit{Shorter experiment} (1500 seconds): A slightly shorter version of the main dataset, at the same walking speed.
 \item \textit{Quad with dynamics} (398 seconds):
 4 clockwise loops around the Quad at a faster walking speed (1.5msec/): for 2 loops the device was held
flat; 1 loop had the device swinging from side to side; and 1 loop where the swinging speed varied.
 \item \textit{Dynamic spinning} (120 seconds): Aggressive angular motions of the device, rotation rates of 2.5 rad/sec while pointing at a corner of the Quad. Useful for testing motion distortion.
 \item \textit{Parkland mound} (500 seconds): Walking a few laps in the grassy area of the Parkland,
up and down the Mound's stairs and finally a loop around the Mound at a fast walking speed.
\end{itemize}

For more details regarding these datasets, we refer the readers to the dataset website.

\section{Ground Truth}
\label{sec:groundTruth}
The ground truth poses of this dataset are obtained with an approach whose core uses Iterative
Closest
Point (ICP), a well-known method for registration of 3D shapes~\cite{besl1992method}. When provided with a
prior pose from which the point clouds are captured, ICP minimizes the Euclidean distance between
closest points, namely correspondences, to iteratively compute the relative displacement between
the reading cloud and the reference cloud. The former refers to the cloud which is intended to be
registered with the latter.

To provide the prior map of our dataset, we use a survey-grade 3D imaging laser scanner,
Leica BLK360\footnote{\label{note3}
https://leica-geosystems.com/en-gb/products/laser-scanners/scanners\\/blk360 }
(\figref{fig:rooster-blk} (bottom)). For the New College
environment with approximately the size of 135$\times$225~m$^2$, 47 static point clouds were captured to fully
map the area. This took over 8 hours. The capture locations were decided to be closer together in the Parkland due to foliage and less
structured features. All the point clouds were matched with over 90$\%$ of inliers. The
fully merged map is seen in~\figref{fig:teaser} (top). According to  the Leica
BLK360 datasheet\footref{note3}], the achievable accuracy for 3D points in the
map is 6 mm at range 10 m and 8 mm at range 20 m. Hence, we can conclude that
the accuracy of the majority of points of the
entire map is better than 1 cm since the range of the points (in the Quad, Mid-Section and the
perimeter of the Parkland)  in the map is no more than 20
meters from at least one of the scanning stations. The final map consists of about 290 million
points.

\begin{figure}[!t]
\centering
 \begin{minipage}{.5\textwidth}
 \centering
   \includegraphics[width=0.9\linewidth]{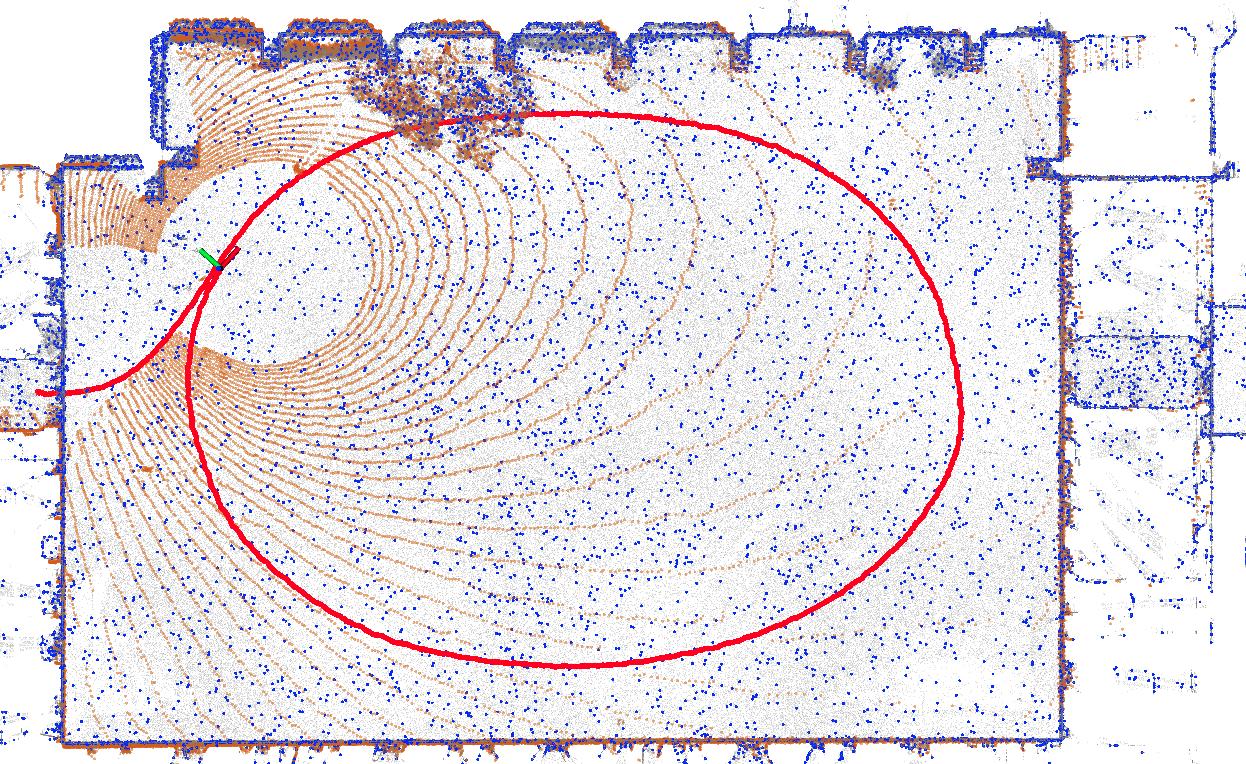}\\
   \vspace{1mm}
  \end{minipage}
  \begin{minipage}{.5\textwidth}
  \centering
   \includegraphics[width=0.9\linewidth]{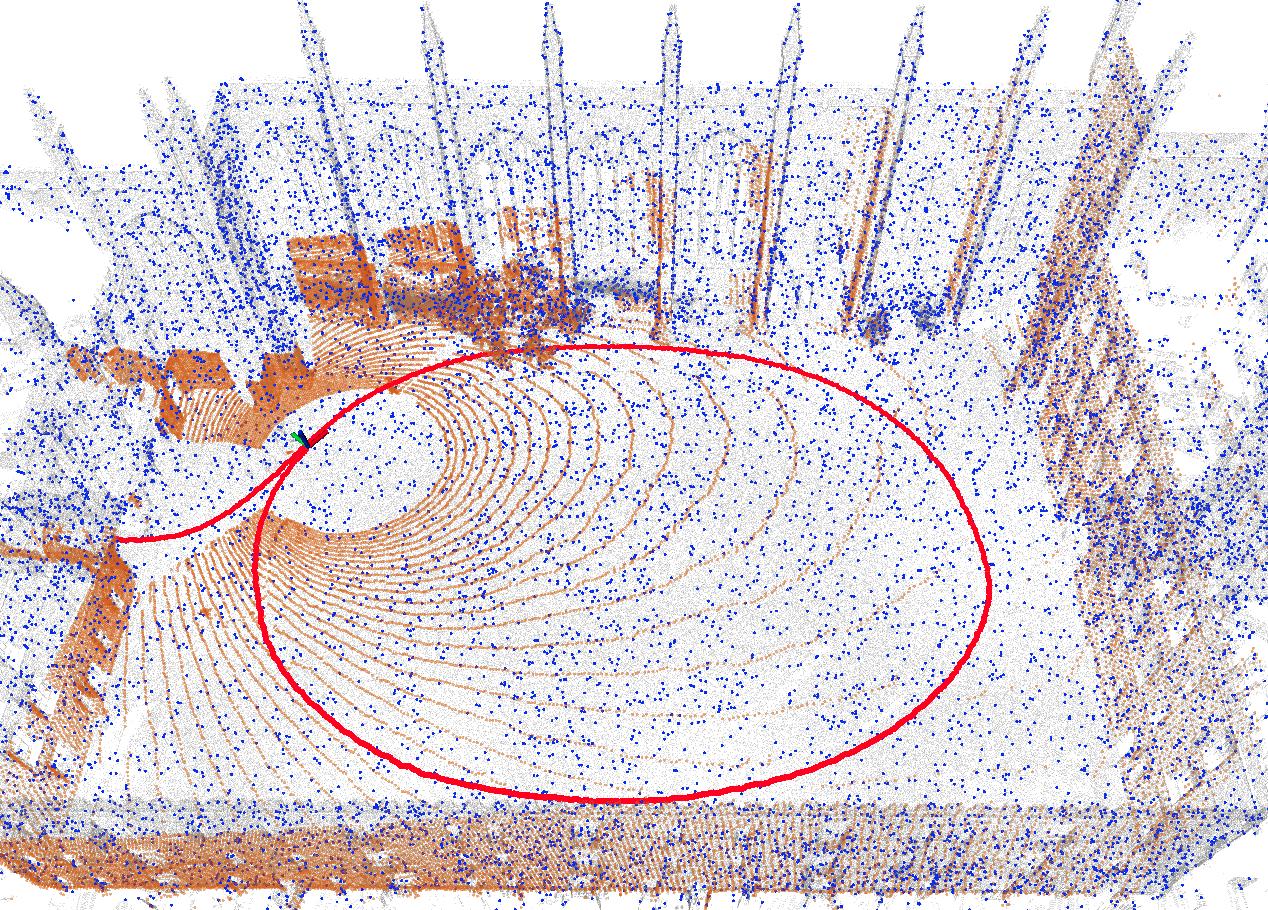}
  \end{minipage}
     \caption{\small{Plan and perspective view of the Quad when current laser scan (in maroon)
     is registered against the reference cloud (in blue). The reference cloud is part
     of the prior map (in gray) which is cropped around the pose.}}
\label{fig:ground-truth}
\end{figure}

Having generated the accurate prior map, we first downsample the map cloud to 1 cm resolution. This
way, we reduce the map to about 17 million points enabling us to use it in our
localisation approach
without an observable drop in registration accuracy. Further, we dynamically crop the pointcloud to
create a
reference cloud in the area of 100 m by 100 m around the sensor's pose. To localise individual
scans, we use a libpointmatcher filter chain~\cite{Pomerleau12comp} to remove the outliers in the
clouds and
finally register the scans against the map.~\figref{fig:ground-truth} demonstrates
the ground truth registration procedure with a single Ouster scan. It is worth noting that the
ground truth poses are
with respect to the base frame which is the center of the left camera, as described
in~\secref{sec:platform}. This facilitates the estimation of the camera poses at 30 Hz through
interpolation.

\begin{figure}[ht]
\centering
\includegraphics[width=1\linewidth,trim={3.5cm 0cm 4cm
1cm},clip]{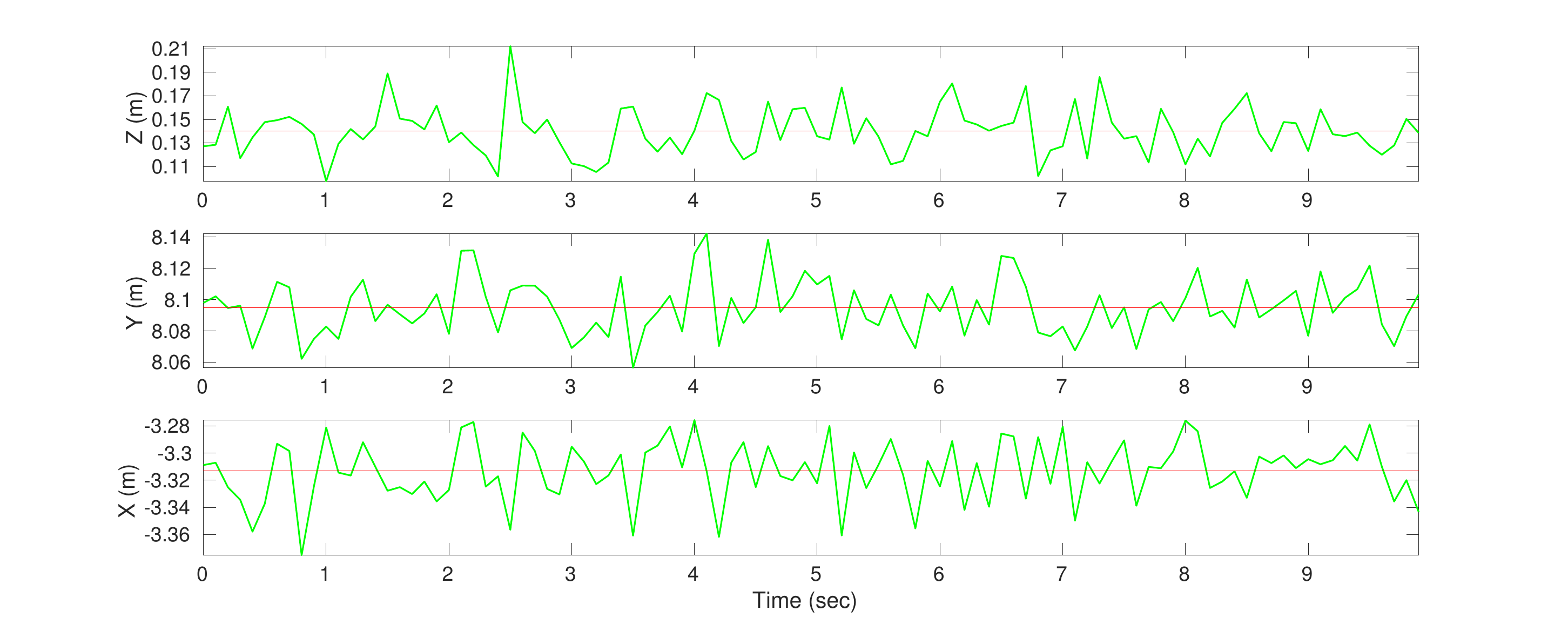}
\caption{\small{Ground truth position values for the first 10 seconds of the dataset when the
device was stationary. Red lines show the mean values over this period of time.}}
\label{fig:gt-stationary}
\vspace{-1mm}
\end{figure}
\begin{figure}[!ht]
\centering
\includegraphics[width=1\linewidth,trim={3.5cm 0cm 4cm
1cm},clip]{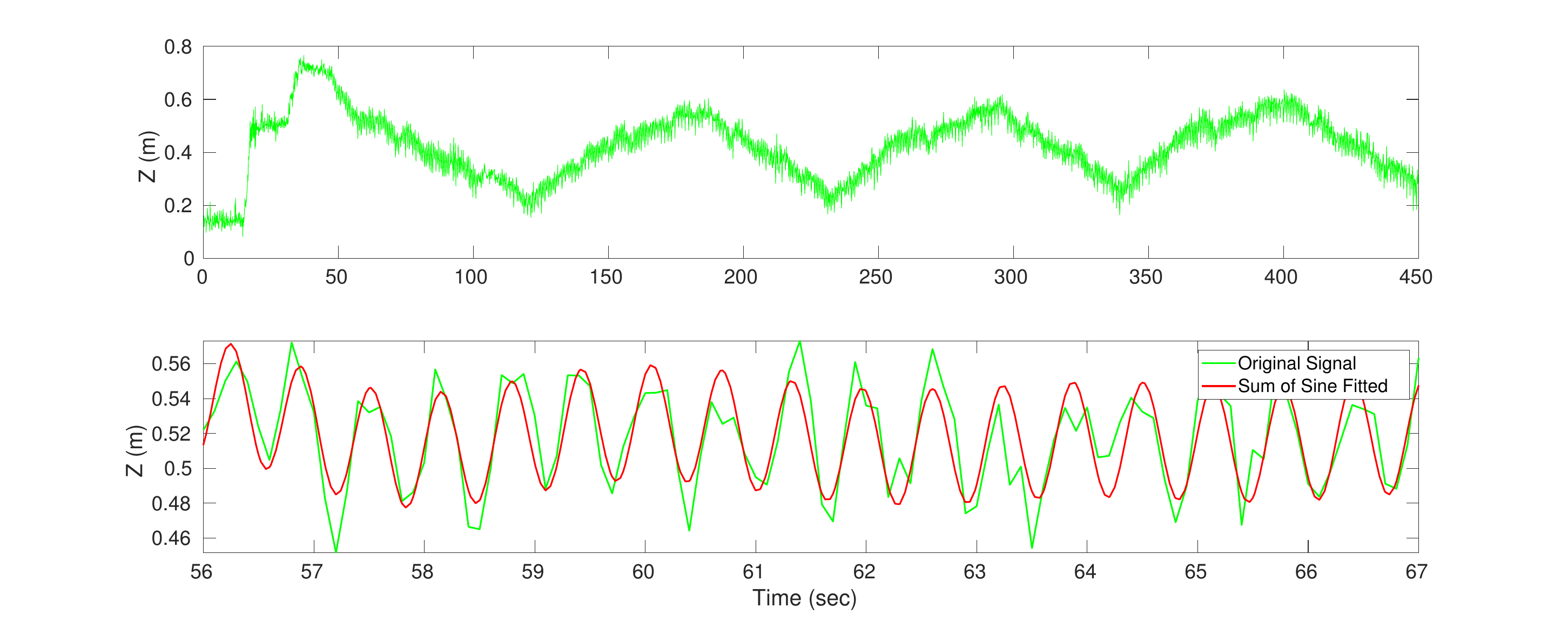}
\caption{\small{\textbf{Top}: Ground truth z values over the course of 450 seconds of the dataset
when
walking around the quad (\textbf{Q1})}. \textbf{Bottom}: A closer analysis over
12 seconds. The fitted red curve shows the pattern of the walking gait.}
\label{fig:gt-eval}
\vspace{-5mm}
\end{figure}

\begin{figure*}[t!]
\centering
\includegraphics[width=1\linewidth,trim={0cm 0cm 0cm
0cm},clip]{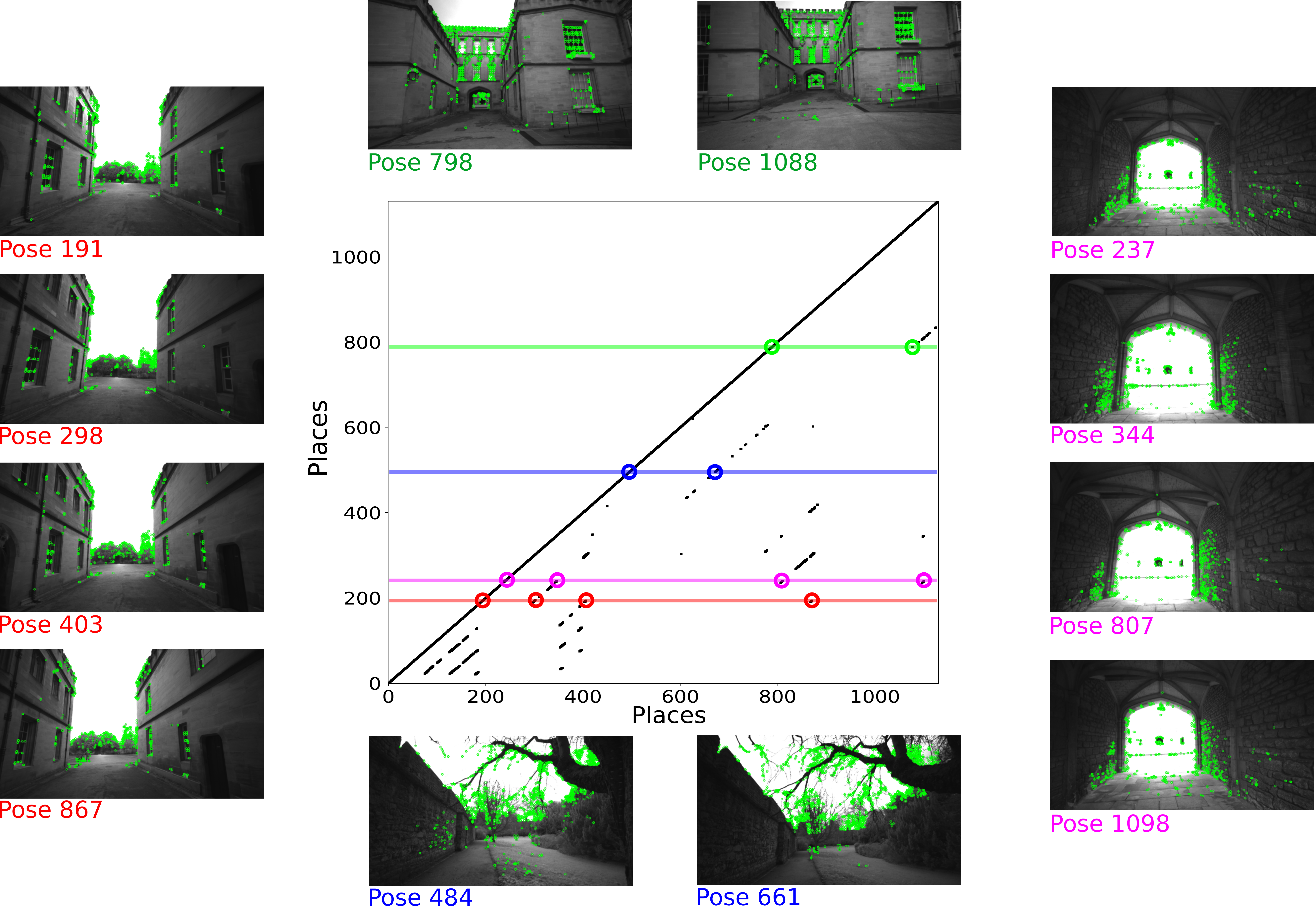}
\caption{\small{An example of dataset usage for appearance-based loop-closure detection.}}
\label{fig:appearance-loopClosures}
\end{figure*}

Evaluation of ground truth accuracy is challenging for certain scenarios in
the dataset -- such as the confined areas in the Mid-section and the natural environment in the
Parkland. \figref{fig:gt-stationary}
and~\figref{fig:gt-eval} illustrate the accuracy of the ground truth.~\figref{fig:gt-stationary}
shows the behaviour of the
ICP ground truth procedure along x, y and z for the first 10 seconds
of the dataset when the device was stationary. The standard deviations along x, y and z are 2 cm,
1.6 cm and 2 cm or
approximately 3 cm overall. Evaluating the localisation performance when the device is in motion is
more challenging.
It involves dynamic effects such as LiDAR motion distortion and it's accuracy likely to be reduced
lower.
Note~\figref{fig:gt-eval} demonstrates the periodic behaviour of the trajectory over the
course of
the three Quad loops (\textbf{Q1}). A closer analysis of 12 seconds of the trajectory
(\figref{fig:gt-eval}, bottom) shows periodicity of the trajectory due to the walking motion --- an
indication
of the accuracy.

\begin{figure*}[t!]
\centering
\begin{multicols}{3}
\resizebox{1\linewidth}{!}{
\includegraphics[width=5cm]{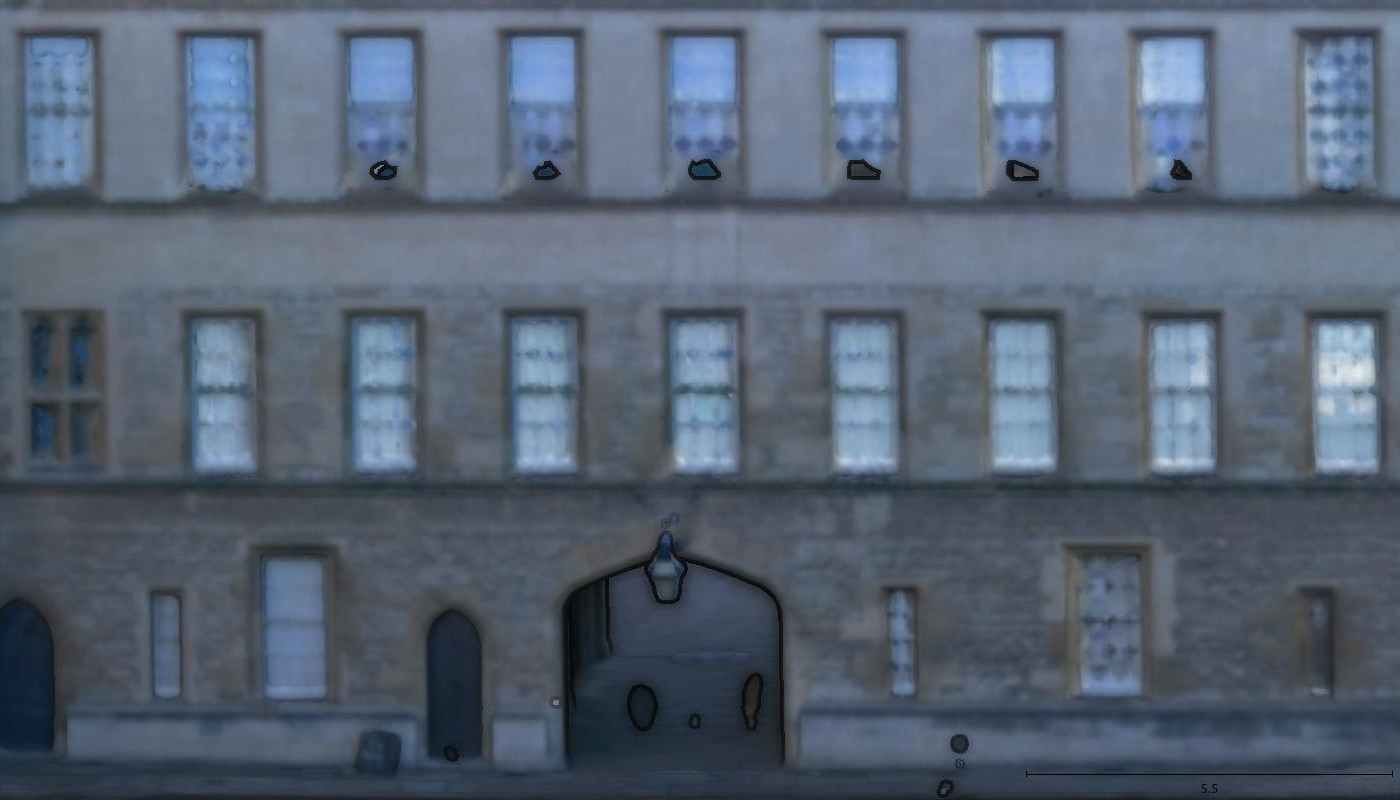}}\vspace{2mm}
\resizebox{1\linewidth}{!}{
\includegraphics[width=5cm]{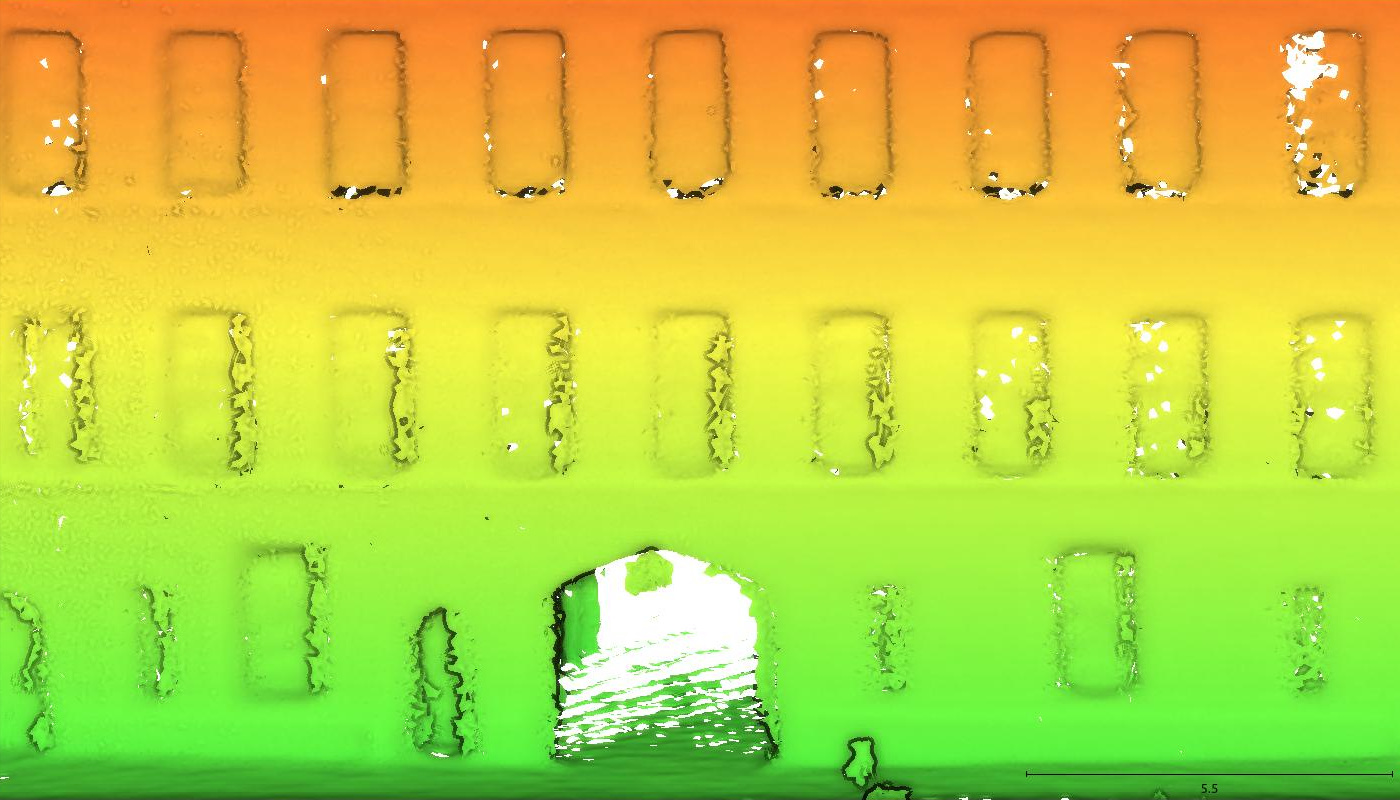}}\\
\resizebox{1\linewidth}{!}{
\includegraphics[width=5cm]{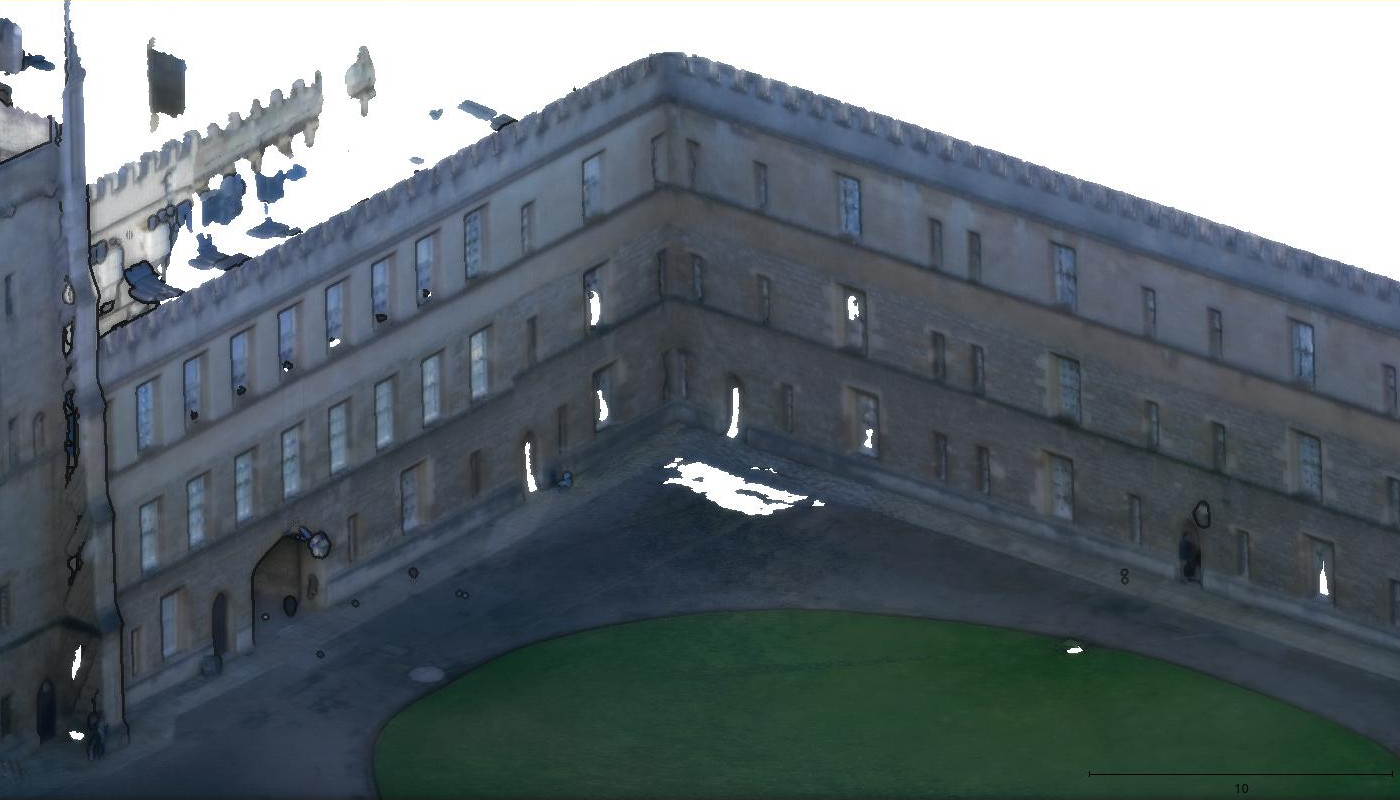}}\vspace{2mm}
\resizebox{1\linewidth}{!}{
\includegraphics[width=5cm]{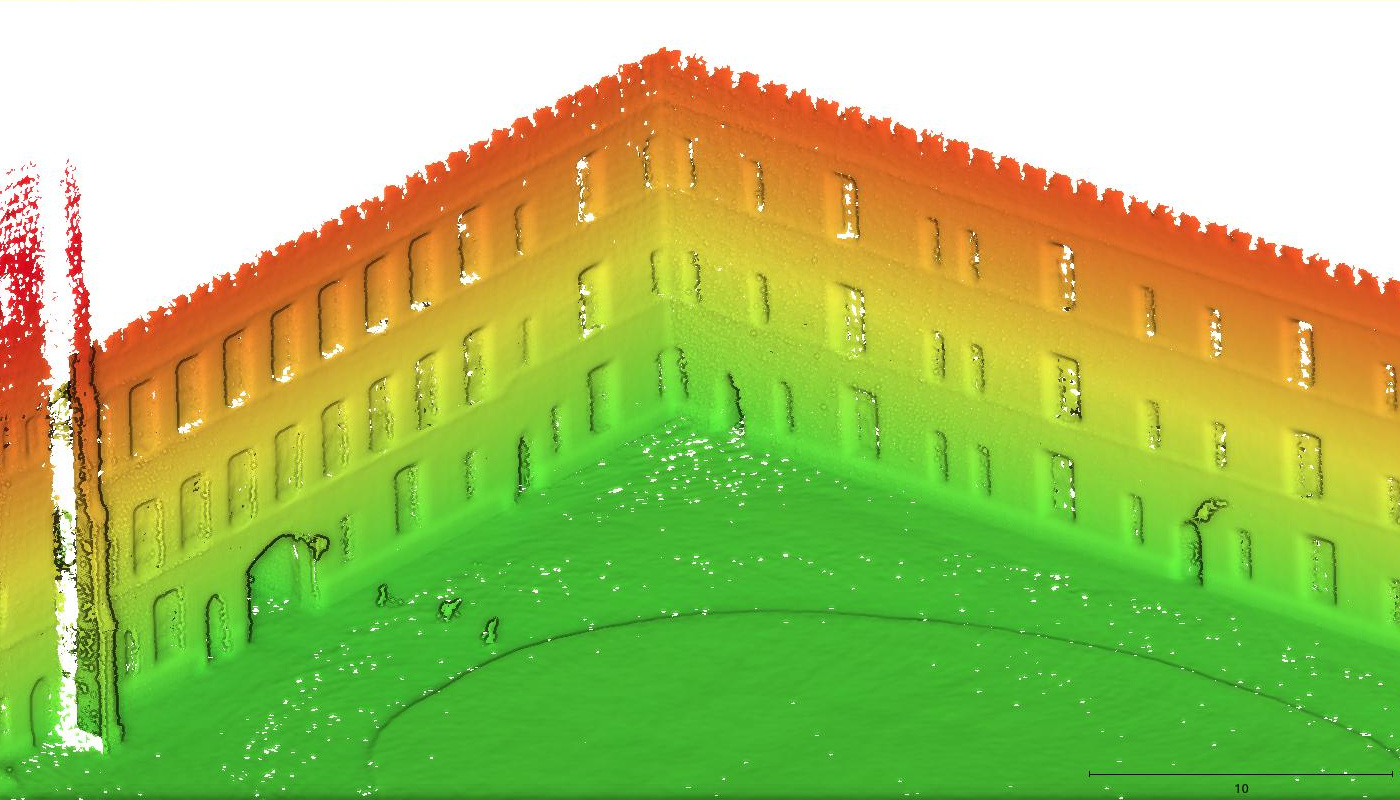}}\\
\resizebox{1\linewidth}{!}{
\includegraphics[width=5cm]{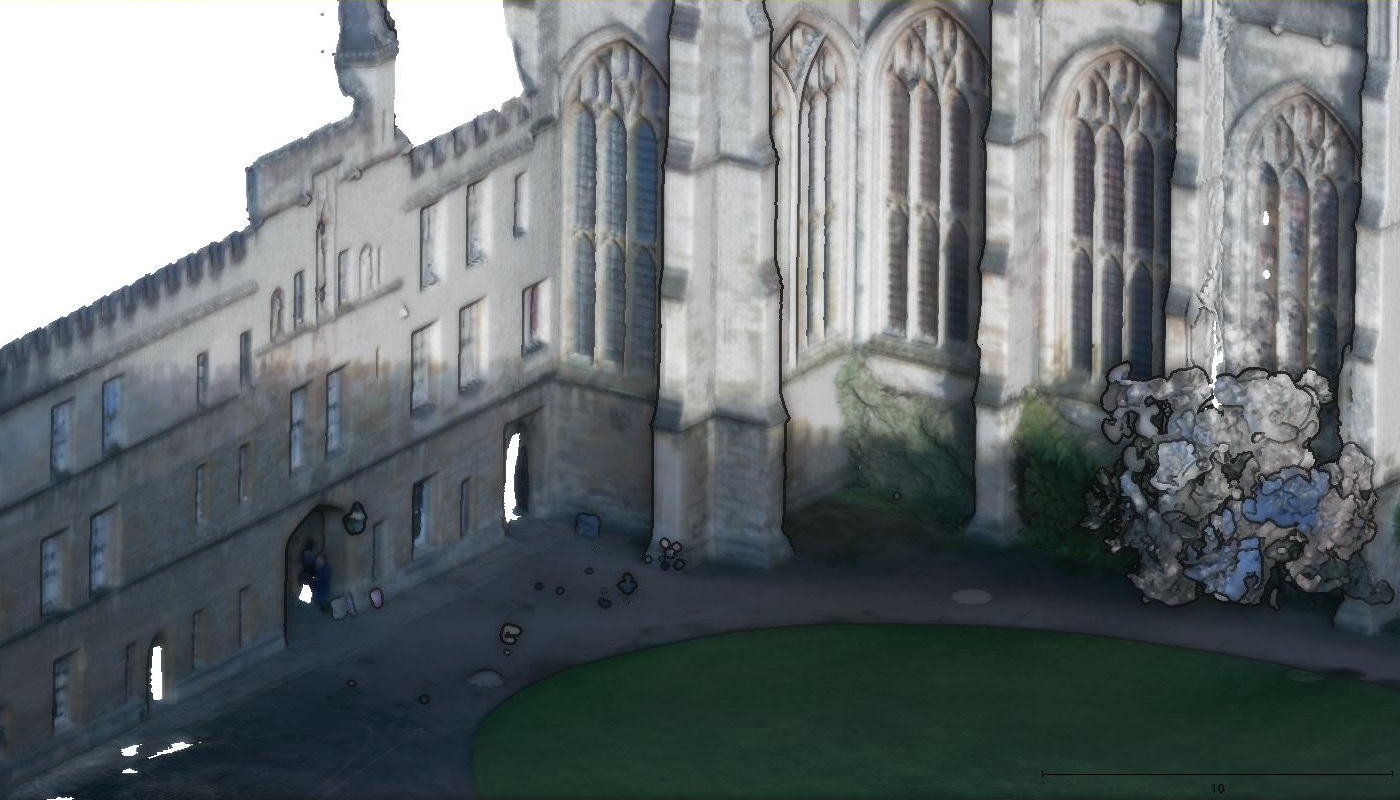}}\vspace{2mm}
\resizebox{1\linewidth}{!}{
\includegraphics[width=5cm]{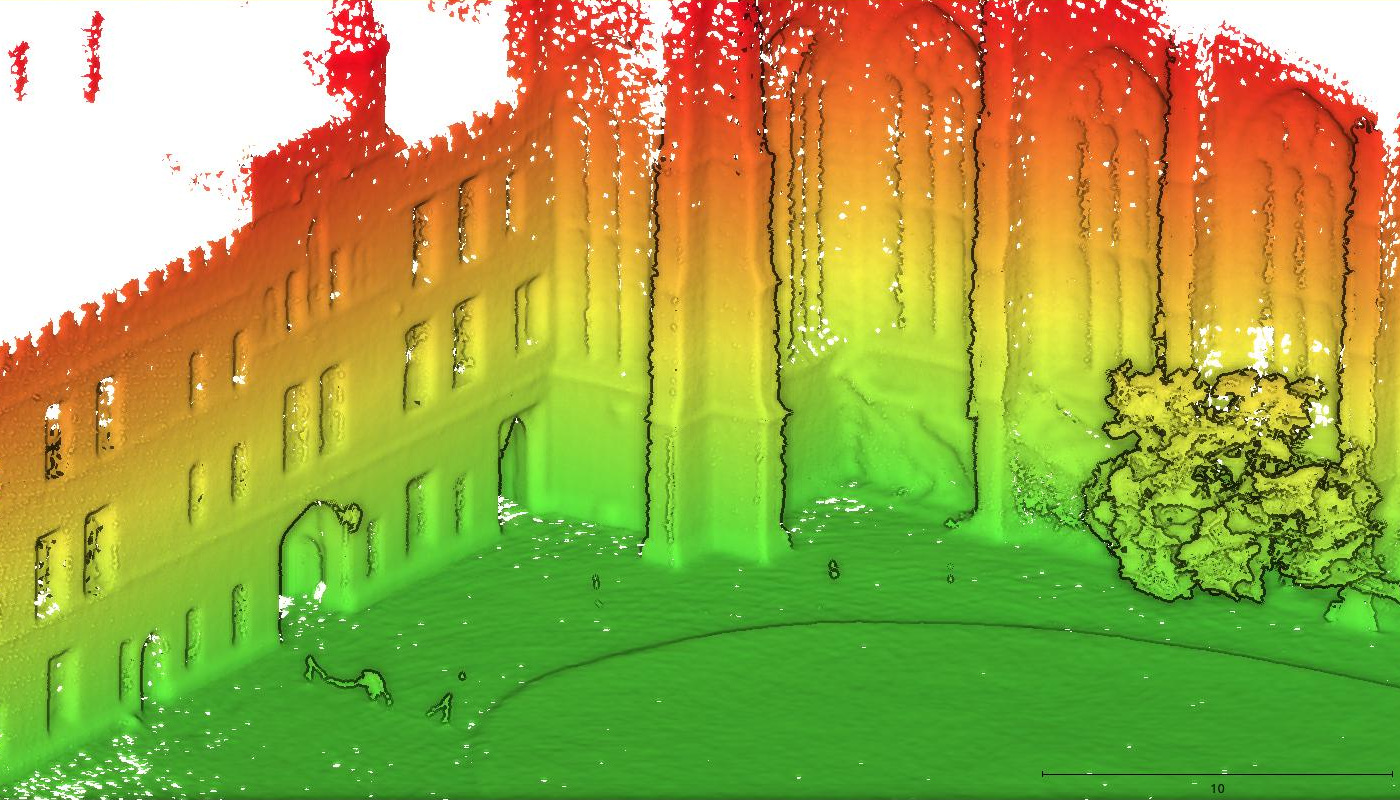}}
\end{multicols}
\vspace{-1em}
\caption{\small{3D reconstruction of the Quad from different view points using Ouster
laser scans (bottom row) and using the ground truth map (top row). The heat map displays the
height from the ground along z axis w.r.t. map frame.}}
\label{fig:3d-reconstruction}
\vspace{-5mm}
\end{figure*}

\begin{figure}[!t]
\centering
 \begin{minipage}{.5\textwidth}
 \centering
   \includegraphics[width=0.9\linewidth,trim={14cm 0cm 13cm 1.5cm},clip]{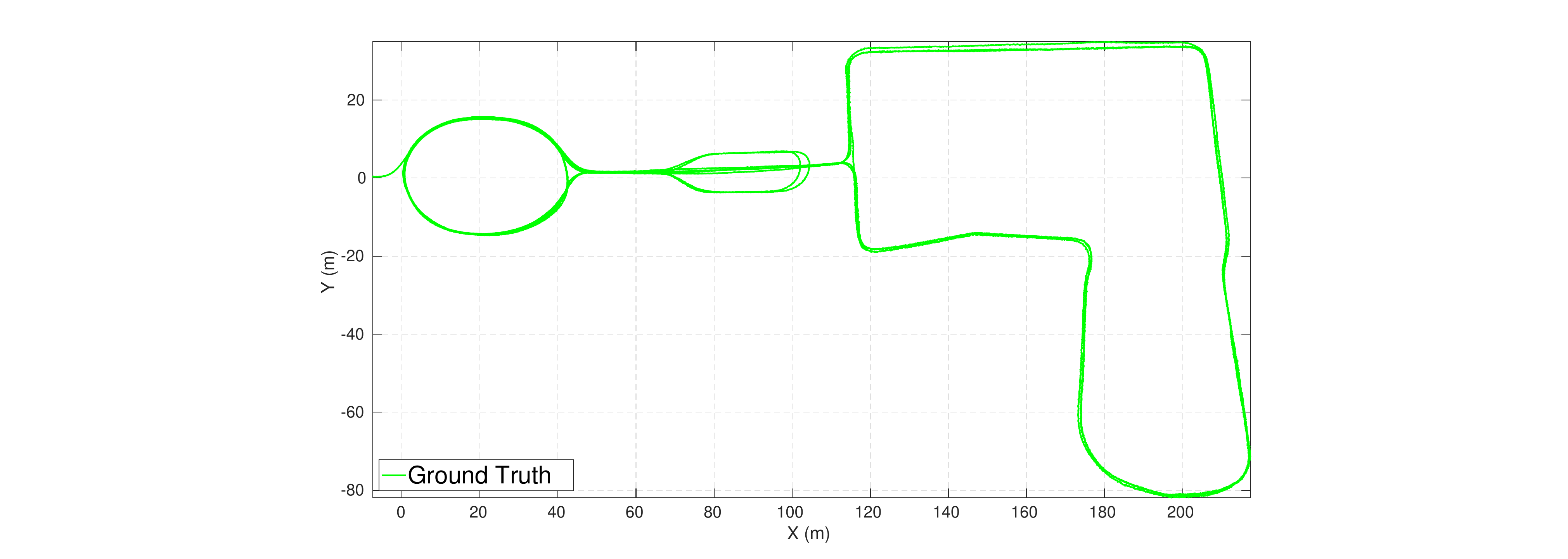}\\
  \end{minipage}
  \begin{minipage}{.5\textwidth}
  \centering
   \includegraphics[width=0.9\linewidth,trim={14cm 0cm 13cm 1.5cm},clip]{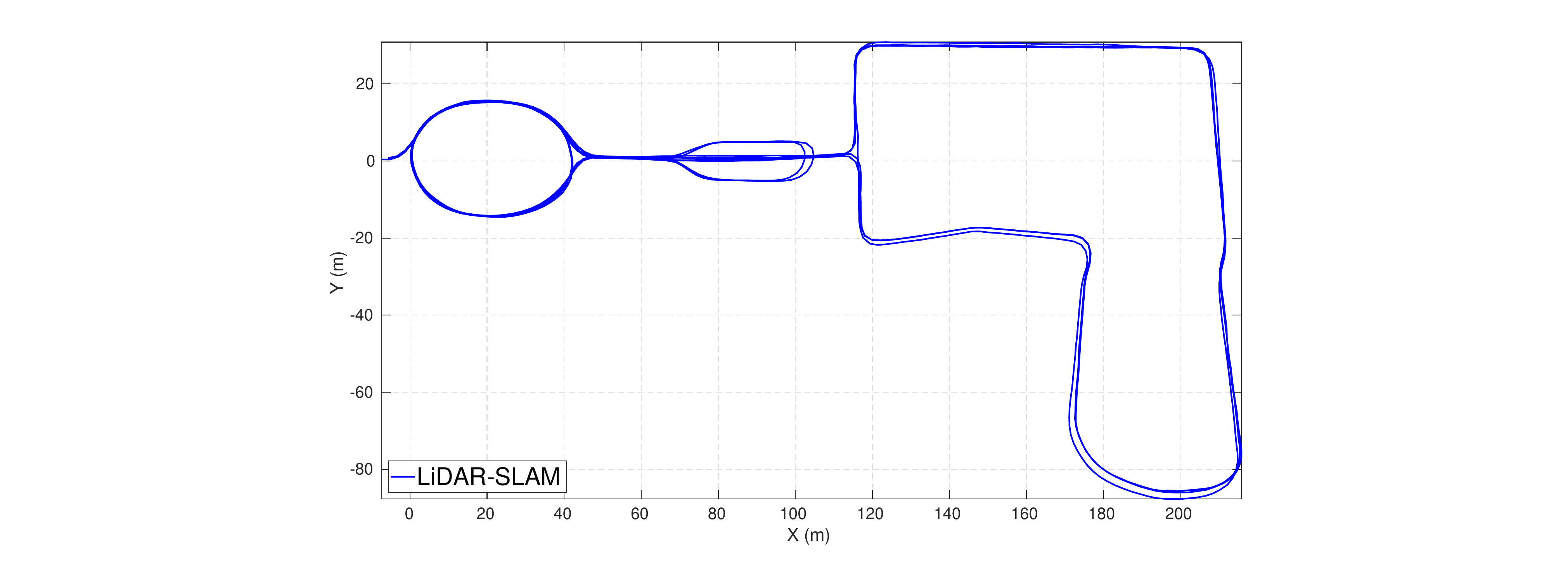}
  \end{minipage}
  \begin{minipage}{.5\textwidth}
  \centering
   \includegraphics[width=0.9\linewidth,trim={19.7cm 0cm 17.6cm
1.5cm},clip]{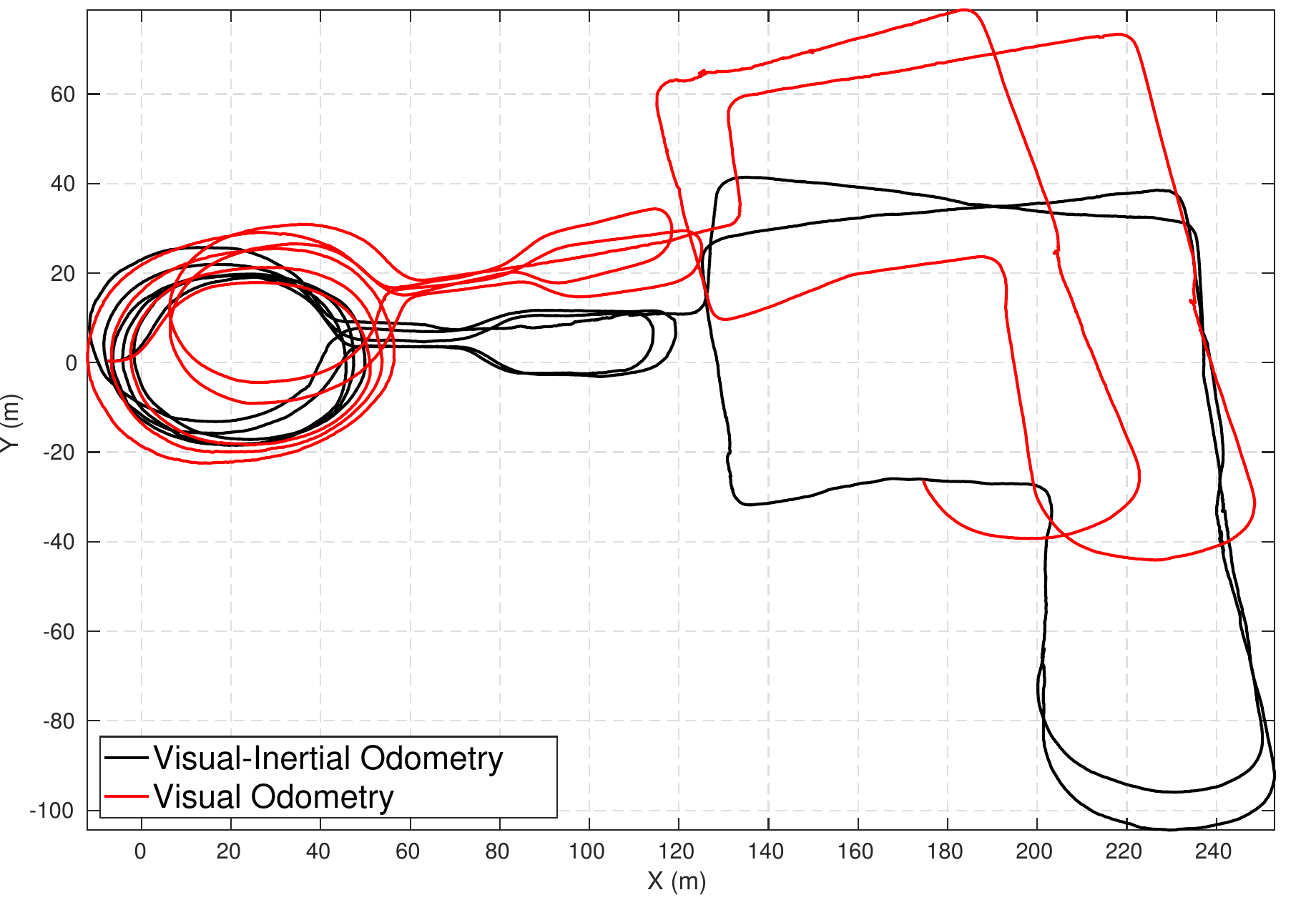}
  \end{minipage}
  \vspace{-2mm}
     \caption{\small{Ground truth trajectory (top) with our LiDAR-SLAM trajectory (middle)
throughout the data. Bottom are the trajectory of vision-based localisation systems, ORB-SLAM2
(red) with loop-closure disabled and VILENS (black).}}
\label{fig:trajs}
\vspace{-5mm}
\end{figure}

\section{Example Dataset Usage}
\label{sec:dataset-usage}
This section presents example results for a series of potential uses of our dataset in localisation and mapping
research areas: LiDAR-SLAM, Appearance-Based
Loop-Closure, 3D Reconstruction and Visual (Inertial) Odometry.

\subsection{LiDAR SLAM}
We use our LiDAR SLAM system detailed in~\cite{ramezani2020online} to estimate ego-motion at
2 Hz and find loop-closures geometrically.~\figref{fig:trajs} (middle) shows the SLAM trajectory for
the entire dataset. This demonstrates that our dataset is useful for LiDAR-based localisation
systems.

\subsection{Visual Appearance-Based Loop-Closure}
As an illustration of our dataset used in visual place recognition, we used
DBoW2~\cite{Galvez-Lopez2012} with
ORB features~\cite{Rublee2011}. We computed the similarity score against all the
poses (spaced at 2 m intervals)
traveled in the past and applied a threshold to obtain loop candidates.~\figref{fig:appearance-loopClosures}
shows the similarity matrix, a square matrix indicates whether the nodes in a pose graph are
similar or not, and some examples of the most similar views captured at different times are shown.

\subsection{LiDAR 3D Reconstruction}
Our dataset can also be used for 3D reconstruction, as demonstrated
in~\figref{fig:3d-reconstruction}. The top row represents the surface mesh generated from the Leica ground truth
map using Poisson surface reconstruction~\cite{kazhdan2006poisson}. The bottom row shows the
mesh created by using the Ouster laser scans registered against the prior map (i.e. leveraging the ground truth
localisation). We applied the
filtering and mesh generation tools provided by PCL~\cite{Rusu_ICRA2011_PCL} including Moving
Least Squares
(MLS) smoothing~\cite{alexa2003computing}, a voxel grid filter of 5~cm resolution and
the greedy triangulation projection~\cite{marton2009fast} for the reconstruction.

\subsection{Visual (Inertial) Odometry}
We provide visual and inertial measurements which are software time synchronized, in our
dataset, as well as providing calibration files to test on two visual odometry methods.
We tested ORB-SLAM2~\cite{mur2017orb} for basic stereo odometry with loop closures disabled.~\figref{fig:trajs} (bottom)
shows the trajectory estimated by this approach using the image pairs for the first 1483 seconds of the
dataset. We also used our visual-inertial odometry approach, VILENS~\cite{wisth2019robust},
which carries out windowed smoothing of the two measurements sources for the same period.

\section{Conclusion and Future Work}
\label{sec:conclusion}
In this paper, we presented the Newer College Vision and LiDAR dataset. By leveraging a 
highly accurate and detailed prior map, we determined accurate 6 DoF ground truth for the entire
dataset, which
distinguishes our dataset from many others. We used a modern visual-inertial camera and
a dense 3D LiDAR sensor and provide the dataset in both the original ROSbags and individual files (such as
png images and csv files).

We also demonstrated the use of the
dataset for different subproblems in mobile robotics and navigation. It is our intent 
to further extend our dataset with additional difficult sequences including
aggressive motions and make them available publicly in
the near future.
A demonstration video is available at:
\url{https://youtu.be/aIeMPeHDUgs}

\section{Acknowledgment}
The authors would like to thank the members of the Oxford Robotics Institute (ORI)
who helped with the creation of this dataset release, especially Chris
Prahacs, Simon Venn and Benoit Casseau. We also thank the personnel of New College
for facilitating our data collection.

This research was supported by the
Innovate UK-funded ORCA Robotics Hub (EP/R026173/1) and the EU H2020 Project THING.
Maurice Fallon is supported by a Royal Society University Research Fellowship. Matias Mattamala is
supported by the National Agency for Research and Development (ANID) / Scholarship Program /
DOCTORADO BECAS CHILE/2019 - 72200291

\bibliographystyle{IEEEtran}
\balance
\bibliography{library.bib}

\end{document}